\documentclass[runningheads]{llncs}
\usepackage[T1]{fontenc}
\usepackage{graphicx}
\usepackage{amsmath}
\usepackage{amssymb}
\begin{document}
\title{Controlled Causal Hallucinations Can Estimate Phantom Nodes in Multiexpert Mixtures of Fuzzy Cognitive Maps}
\titlerunning{Phantom Node Estimation in FCM Mixtures}
%
\author{Akash Kumar Panda \and
Bart Kosko}
%
%
\institute{
Signal and Image Processing Institute\\
Department of Electrical and Computer Engineering\\
University of Southern California\\
\email{kosko@usc.edu}}
\maketitle              
\vspace{-12pt}
\begin{abstract}
An adaptive multiexpert mixture of feedback causal models can approximate missing or phantom nodes in large-scale causal models.
The result gives a scalable form of \emph{big knowledge}.
The mixed model approximates a sampled dynamical system by approximating its main limit-cycle equilibria.
Each expert first draws a fuzzy cognitive map (FCM) with at least one missing causal node or variable.
FCMs are directed signed partial-causality cyclic graphs.
They mix naturally through convex combination to produce a new causal feedback FCM. 
Supervised learning helps each expert FCM estimate its phantom node by comparing the FCM's partial equilibrium with the complete multi-node equilibrium.
Such phantom-node estimation allows partial control over these causal hallucinations and helps approximate the future trajectory of the dynamical system.
But the approximation can be computationally heavy.
Mixing the tuned expert FCMs gives a practical way to find several phantom nodes and thereby better approximate the feedback system's true equilibrium behavior. 

\end{abstract}
\section{Mixing Expert Causal Feedback Models to Find Phantom Causal Nodes}
\vspace{-1pt}
Building large-scale causal models faces a key epistemic problem:  \emph{What are the relevant but \emph{missing} causal variables in a causal model}? 
Where do these missing nodes come from?  
Who or what neural or other data-mining system comes up with them?

This paper gives a working answer for the hard causal problem of how to model a sampled multivariable feedback dynamical system $\dot{x} = f(x)$ that has multiple variables.
The technique uses a large number of experts or other knowledge sources.

The answer has two parts:  Experts first guess at the relevant nodes that affect some equilibrium behavior.
But they allow that there is at least one missing node that they should include.  
Figure 1 shows this first step for a single expert where the expert can be a human or neural network or any other knowledge source or data-based algorithm.
Then we mix or combine these causal models in a high-level probability mixture of the causal models and then tune from there.
Figure 2 shows this second step for three experts whose causal mixture estimates three distinct missing or phantom causal nodes.

This multiexpert adaptive process gives a causal mixture of experts that is itself a \emph{feedback} dynamical system that approximates the sampled underlying dynamical system.
The express use of feedback or cyclic models rules out using acyclic DAG causal models \cite{ogburn2024causal,castelletti2020discovering,pearl2009causality}.
DAGs are feedforward systems and so have no nontrivial dynamical equilibria because they have no dynamics at all.
We focus in contrast on estimating discrete equilibrium limit cycles.

\begin{figure}[ht]
\centering
\includegraphics[width=\textwidth]{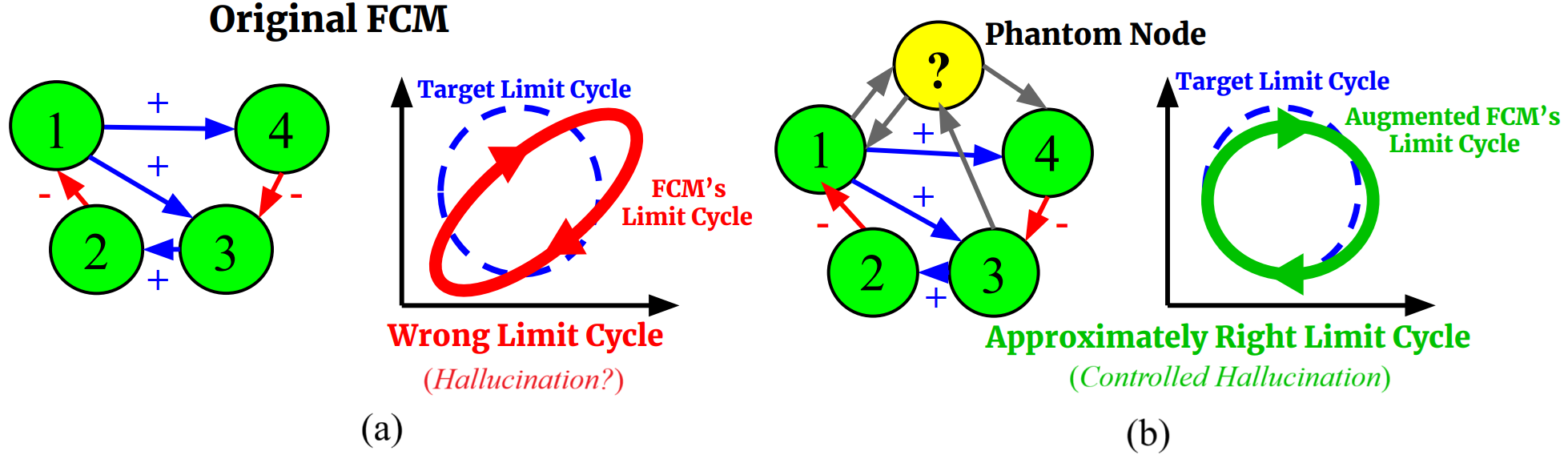}
\caption{Causal phantom nodes augment a FCM and help approximate the FCM dynamics and its target equilibrium limit cycles. 
The figures on the left show the FCMs and the figures on the right show their corresponding limit cycle. 
(a) The original FCM with 4 nodes. 
(b) The augmented FCM with 4 observable nodes and one phantom node. 
The original FCM does not approximate the limit cycles of the system it models. 
The FCM with the phantom node can approximate the target limit cycles with its own limit cycles.}
\label{fig:phantom-node}
\end{figure}

\begin{figure*}[ht]
\centering
\includegraphics[width=\textwidth]{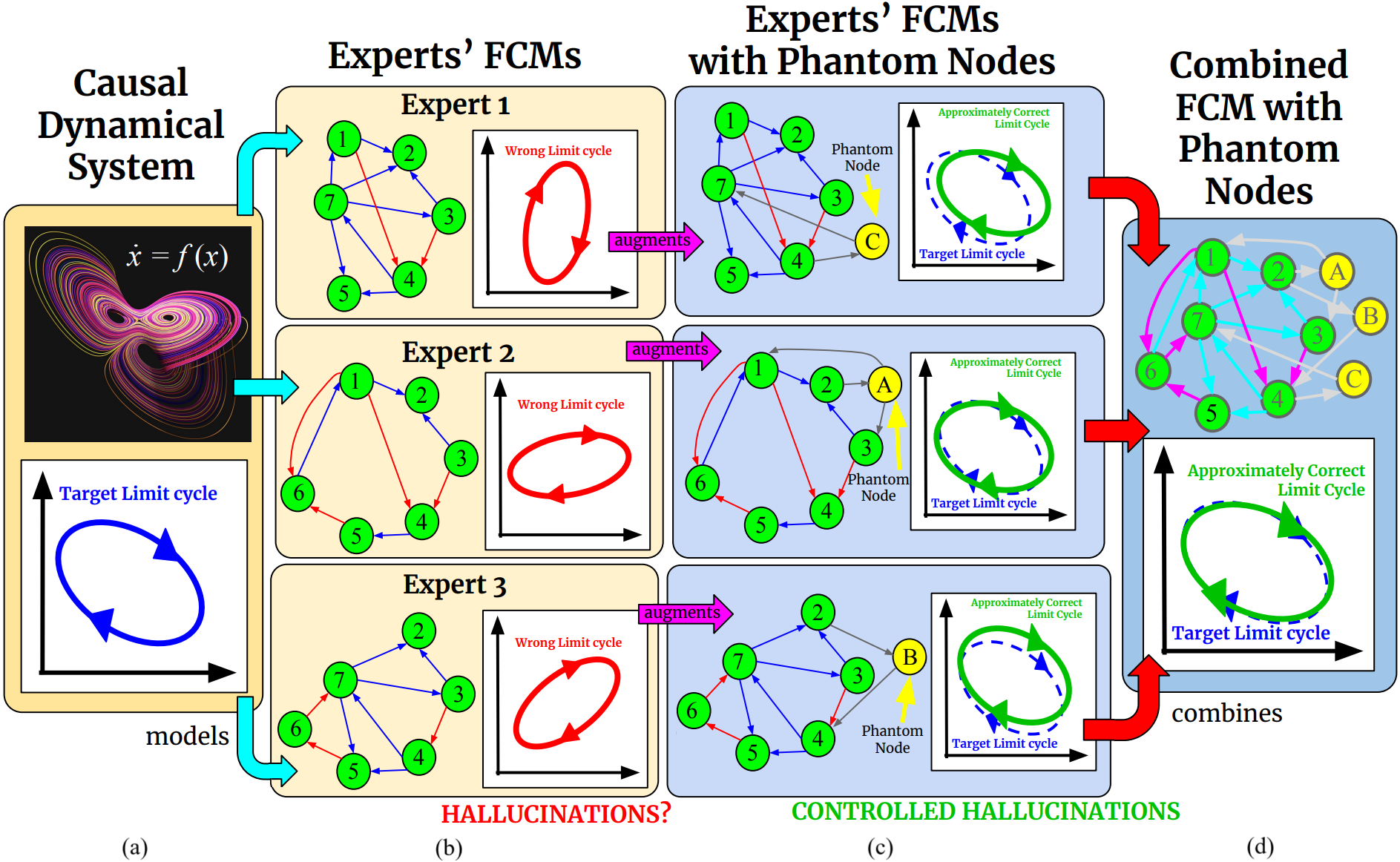}
\caption{Phantom nodes augment experts' FCMs that then combine to approximate the limit cycles of a dynamical system through controlled hallucinations. 
(a) Causal Dynamical system that the expert FCMs model and its target limit cycle that the FCMs try to approximate. 
(b) Three 6-node FCMs from 3 different experts and their corresponding limit cycles.
The limit cycles do not match those of the dynamical system. 
This indicates the presence of phantom nodes. 
(c) Phantom nodes $A$, $B$, and $C$ respectively augment FCMs from experts 2, 3, and 1. 
Phantom nodes allow controlled hallucinations and approximate the dynamical system's limit cycle. 
(d) The 3 7-node FCMs mix into a 10-node FCM through convex combination. 
The combined FCM hallucinates a limit cycle even closer to that of the dynamical system. }
\label{fig:controlled-hallucination}
\end{figure*}

The missing nodes or variables are \emph{phantom nodes} if we analogize them to the well-documented neural hallucination of missing phantom limbs \cite{ramachandran2000phantom}.
Detecting phantom nodes involves what we can view as a rough form of causal or AI hallucination \cite{alkaissi2023artificial}.
The induced creative or hallucinatory outcome resembles adding noise to a nonlinear system in stochastic resonance \cite{kosko2003stochastic} where a small amount of noise can improve system performance but too much noise can harm it. 
We have also found with phantom nodes that a small amount of \emph{causal hallucination} or noise-like perturbation can help estimate a missing relevant causal node or variable while too much hallucination tends to obscure it. 
We can better approximate the dynamical system if we control these causal hallucinations. 

We use feedback fuzzy cognitive maps (FCMs) for this difficult estimation task.
FCMs model causality in complex feedback dynamical systems \cite{kosko1986differential,kosko1986fuzzy,kosko1988hidden,osoba2017fuzzy,ziv2018potential,glykas2010fuzzy,papageorgiou2013fuzzy,stach2010divide,kosko1988hidden,taber2007quantization,apostolopoulos2024fuzzy}.
FCMs are signed directed cyclic graphs of partial or fuzzy causality where a positive sign indicates causal increase while a negative sign indicates causal decrease.  
FCMs naturally combine by mixing their partial-causality edge matrices in a weighted average.
Missing nodes correspond to zero-padded rows and columns in an augmented edge matrix.

Even detecting a single causal phantom can be computationally heavy.
A missing or phantom node can causally affect the other variables in a causal model and further do so in still more complex ways in a final mixture model.
These changes affect causal nonlinear system's dynamics and thus can easily change its equilibrium causal predictions \cite{hoyer2008estimation,ito2016backward,chen2020mining}. 

These changes can also change the dynamics of a feedback nonlinear FCM and thereby change its equilibrium attractors such as its fixed points or limit cycles.
These attractors act as answers.
They are how the FCM's what-if predictions given input stimuli. 
This change in dynamics and equilibrium structure can also suggest the presence and nature of the phantom variables.
That is the idea behind the limit-cycle-based algorithm in this paper that estimates the causal structure of a phantom node from known target limit cycles.
The known limit cycles allow validation and supervised learning of the causal edges to and from the phantom concept node.
Such limit cycles may correspond in practice to a user's desired policy outcome.

Section~\ref{FCM} gives the details of simple feedback FCMs. 
Section~\ref{Causality} explains how FCMs model causality as weighted signed directed graphs. 
Figure~\ref{fig:dolphin-fcm} shows an example of a FCM that models behavior in an undersea herd of dolphins. 
Section~\ref{FCM-update} describes the time evolution of a FCM. 
It explains how FCMs go from their current state to the next state using matrix multiplication and thresholding. 
Section~\ref{FCM-equilibria} describes different types of equilibrium behaviors of a FCM. 
Figure~\ref{fig:threshold-limit-cycles} gives examples of dolphin FCM limit cycles. 
Section~\ref{FCM-combination} explains how FCMs can mix through convex combination in contrast to Directed Acyclic Graphs (DAGs).

Section~\ref{Phantom Nodes} explains the concept of phantom nodes. 
Figure~\ref{fig:phantom-dolphin-fcm} gives an example of a phantom node in the dolphin FCM. 
Section~\ref{Phantom-effect} describes the effect that phantom nodes can have on the FCM's limit cycles. 
Figure~\ref{fig:phantom-effect} shows an example of dolphin limit cycles in the presence and the absence of the phantom node ``{\footnotesize SURVIVAL~THREAT''. 
Section~\ref{Phantom-Estimation} gives a way of estimating phantom nodes by comparing the FCM's limit cycles to that of the underlying dynamical system. 
Figure~\ref{fig:approx-limit-cycles} shows the approximate limit cycles from the phantom-node-augmented dolphin FCM. 

Section~\ref{phantom-combination} describes the process of combining multiple FCMs each augmented with phantom nodes. 
Figure~\ref{fig:controlled-hallucination} shows this process. 

Section~\ref{experiments} gives the results from our experiments. 
Figures~\ref{fig:Augmented-FCM-Mixture-limit-cycles} and \ref{fig:Augmented-FCM-wrong-limit-cycles} show that the mixture of augmented FCMs performs better than do its component FCMs.

\section{Fuzzy Cognitive Maps}\label{FCM}
\vspace{-6pt}
Fuzzy Cognitive Maps (FCMs) model causal behavior in a dynamical system as a directed weighted \emph{cyclic} graph. 
The directed edges describe the causal relationships between concept nodes. 
FCMs allow feedback and therefore converge to non-trivial equilibria like limit cycles. 
FCMs can model a dynamical system by approximating its limit cycles. 
They take the current state of the dynamical system as input and give the future trajectory of the system as output. 

The FCMs allow causal cycles whereas Directed Acyclic Graphs (DAGs) do not.
This lets us mix FCMs together through convex combination. 
DAGs cannot mix in general because a mixture of two or more acyclic graphs tends to have cycles. 
FCM mixing is closed:  Mixing FCMS always gives back a FCM. 
The mixed FCM often has a much richer causal and dynamical structure than do the mixed FCMs.
So in this and in other senses FCM mixtures tend to improve with expert sample size.

\vspace{-6pt}
\subsection{Causal Modelling through FCMs} \label{Causality}

The direct edges of the FCMs describe the causal relation between the concept nodes in the FCM. 
An edge $e_{ij}$ from $C_i$ to $C_j$ says "$Ci$ causes $C_j$". 
The weight of the edge $e_{ij}$ describes the degree to which $C_i$ causes $C_j$. 
\begin{align}
    e_{ij} = Degree(C_i \rightarrow C_j)
\end{align}\label{eq:eij}
FCMs allow ``partial causality". 
The weight $e_{ij} \in [-1, 1]$. 
A positive $e_{ij}$ means that an increase in $C_i$ causes an increase in $C_j$ and a negative $e_{ij}$ means that an increase in $C_i$ causes a decrease in $C_j$. 
A high magnitude of $e_{ij}$ means $C_i$ ``strongly" causes $C_j$ and $e_{ij}$ close to 0 means $C_i$ ``weakly" causes $C_j$. 
There is no causal relationship between $C_i$ and $C_j$ if $e_{ij} = 0$ or if there is no edge between $C_i$ and $C_j$. 

A $n\times n$ matrix $E$ can represent the edges of the FCM. 
The weight $e_{ij}$ for the edge connecting the node $C_i$ to the node $C_j$ corresponds to the matrix element on the $i$th row and the $j$th column. 
The concept-node pairs with no edge connecting them correspond to 0 values in the edge matrix. 

Consider the 5-node FCM in Figure~\ref{fig:dolphin-fcm}. 
The FCM describes the group behavior of a dolphin pod in the presence of a survival threat such as a shark. 
The causal-edge matrix $E$ for this dolphin FCM is
\begin{align}
E =     \begin{pmatrix}
0 & 1 & 0 & -1 & 0\\
0 & 0 & 1 & 0 & -1\\
0 & -1 & 0 & 1 & -1\\
1 & 0 & -1 & 0 & 1\\
-1 & 1 & 0 & -1 & 0    \label{eq:fcm-edge-matrix}
\end{pmatrix}.
\end{align}
The edge weight $e_{45}$ from node $C_4$ to $C_5$ is 1 because the presence of a ``{\footnotesize SURVIVAL THREAT}" like a shark causes the dolphins to ``{\footnotesize RUN AWAY}". 
The edge weight $e_{54}$ similarly is -1 because running away gets the dolphins away from the shark. 
\begin{figure}[ht]
\centering
\includegraphics[width=0.55\textwidth]{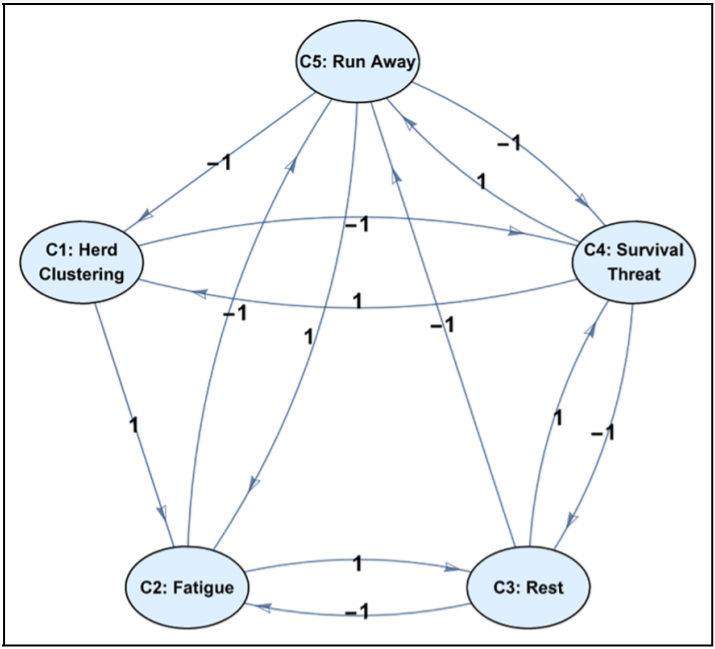}
\caption{Dolphin FCM of dolphins near a threat such as a shark. }
\label{fig:dolphin-fcm}
\end{figure}

\vspace{-6pt}
\subsection{FCM State Evolution} \label{FCM-update}

A $n$-dimensional vector $C(t) \in [0,1]^n$ can represent the state of a $n$-node FCM at time $t$. 
The $i$th concept node is ``active" at time $t$ if $C_i(t)$ is equal to or close to 1. 
The $i$th concept node is ``inactive" at time $t$ if $C_i(t)$ is equal to or close to 0. 
The node is ``partially active" otherwise. 
An active node indicates the presence of the corresponding causal factor in the system and an inactive node indicates the absence of the corresponding causal factor. 

The state $C(t+1)$ of the $j$th node at time $t+1$ is
\begin{align}
    C_j(t+1) = \Phi\bigg(\sum_{i=1}^n  C_i(t)e_{ij}\bigg)\label{eq:next-state}
\end{align}
where $\Phi$ is an increasing nonlinear threshold-like function such that $0\leq\Phi(x)\leq1$. 
The sum inside the parentheses in equation~\ref{eq:next-state}is the matrix product between the state row-vector $C(t)$ and the edge matrix $E$. 

Consider the dolphins at ``{\footnotesize REST}" in the presence of a shark. 
The product of the state vector $C(t) = \begin{pmatrix}0&0&1&0&0\end{pmatrix}$ and the edge matrix $E$ gives $\begin{pmatrix}1&0&-1&0&1\end{pmatrix}$ that thresholds to the state vector $C(t+1) = \begin{pmatrix}1&0&0&0&1\end{pmatrix}$. 
This means the dolphins run away in herd clusters. 

This process repeats to give the time-evolution of the FCM. 
The FCM will start at an initial state $C(0)$ and then go through the states $C(1)$, $C(2)$, $C(3)$, and so on in order. 
This gives the evolution of the system that the FCM models over time. 

A node can also clamp to a constant value $c$. 
The clamped node $C_j(t)$ equals $c$ for all time $t$ no matter what Equation~\ref{eq:next-state} says it should equal.

\vspace{-6pt}
\subsection{FCM Equilibria} \label{FCM-equilibria}

A sequence of $n$-dimensional state vectors $C(t)$ describe the dynamics of a FCM. 
The FCM can have different equilibrium behaviors depending on how state-vector sequence turns out. 
The FCM converges to a fixed point if the state vector converges to a constant vector. 
The FCM converges to a $K$-step limit cycle if there is $C(t+K) = C(t)$ in the state-vector sequence. 
Then a set of $K$-state vectors will repeat themselves over and over in the same order. 
The FCM may converge to a chaotic attractor if there is no repeating pattern in the state-vector sequence. 

The set of initial states that lead to a given equilibrium attractor describe the basin of attraction for said equilibrium attractor. 
These basins of attractions partition the FCM's state space. 
This map from the basins to the attractors characterizes the FCM and the dynamical system it models. 

Figure~\ref{fig:threshold-limit-cycles} gives 3 examples for limit cycles in the dolphin FCM.
The initial states $\begin{pmatrix}0&0&0&1&0\end{pmatrix}$ and $\begin{pmatrix}0&1&0&1&0\end{pmatrix}$ respectively in the first and third example lead to the same limit cycle ``{\footnotesize SURVIVAL THREAT}"$\rightarrow$ ``{\footnotesize RUN~AWAY}" \& ``{\footnotesize HERD~CLUSTERING}"$\rightarrow$ ``{\footnotesize FATIGUE}"$\rightarrow$ ``{\footnotesize REST}"$\rightarrow$ ``{\footnotesize SURVIVAL~THREAT}". 
So both $\begin{pmatrix}0&0&0&1&0\end{pmatrix}$ and $\begin{pmatrix}0&1&0&1&0\end{pmatrix}$ lie in the same basin of attraction.

\begin{figure}[ht]
\centering
\includegraphics[width=0.7\textwidth]{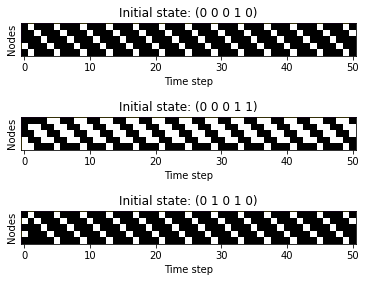}
\caption{Limit cycles from the Dolphin FCM. 
The time step is along the $x$-axis. 
The nodes are along the $y$-axis. 
The images have 5 rows of pixels because the FCM has 5 nodes and each row represents the time evolution of one node.
The color of the node represents its value. 
A bright color corresponds to a high value. 
White nodes have value 1. 
Black nodes have value 0. 
The top figure starts with the initial state $\begin{pmatrix} 0 & 0 & 0 & 1 & 0 \end{pmatrix}$ and the bottom figure starts with the initial state $\begin{pmatrix} 0 & 1 & 0 & 1 & 0 \end{pmatrix}$. They both fall into the same limit cycle: $\begin{pmatrix}0 & 0 & 0 & 1 & 0\end{pmatrix}\rightarrow\begin{pmatrix}1 & 0 & 0 & 0 & 1\end{pmatrix}\rightarrow\begin{pmatrix}0 & 1 & 0 & 0 & 0\end{pmatrix}\rightarrow\begin{pmatrix}0 & 0 & 1 & 0 & 0\end{pmatrix}\rightarrow\begin{pmatrix}0 & 0 & 0 & 1 & 0\end{pmatrix}$. 
The middle figure starts with initial state $\begin{pmatrix} 0 & 0 & 0 & 1 & 1 \end{pmatrix}$ and falls into a different limit cycle: $\begin{pmatrix}1 & 0 & 0 & 1 & 0\end{pmatrix}\rightarrow\begin{pmatrix}1 & 1 & 0 & 0 & 1\end{pmatrix}\rightarrow\begin{pmatrix}0 & 1 & 1 & 0 & 0\end{pmatrix}\rightarrow\begin{pmatrix}0 & 0 & 1 & 1 & 0\end{pmatrix}\rightarrow\begin{pmatrix}1 & 0 & 0 & 1 & 0\end{pmatrix}$.}
\label{fig:threshold-limit-cycles}
\end{figure}

\vspace{-6pt}
\subsection{FCM Combination} \label{FCM-combination}

FCMs mix through convex combination. 
Consider $m$ FCMs. 
Let $S_i$ denote the set of nodes from the $i$th FCM with edge matrix $E_i$. 
The node-set $S$ for the combined $N$-node FCM is the union $S_1\cup S_2\cup S_3\cup\dots\cup S_m$. 
The $N\times N$ matrix $\Tilde{E_i}$ pads $E_i$ with zero-rows and zero-columns corresponding to nodes in $S-S_i$. 
The edge matrix $E$ for the combined FCM is 
\begin{align}
    E = \sum_{i=1}^m w_i\Tilde{E_i} \label{eq:fcm-combination}
\end{align}
where the mixing weights $w_i$ are convex weights such that $w_i\geq0$ and $\sum_{i=1}^m w_i = 1$. 

Then mixing causal edge matrices is closed:  Mixed FCMs always produce an FCM because the mixed edge values remain in the bipolar interval $[-1, 1]$ if the original edge values were bipolar.
This crucially holds when mixing or combining \emph{augmented} edge matrices as we illustrate below.

FCM mixing allows FCMs to combine knowledge from multiple experts because it allows experts to build construct causal models using different or overlapping concept nodes.
The experts construct FCMs based on their understanding of the dynamical system and the causal variables involved in it. 
We can then combine their weighted FCMs where the weights can reflect expert credibility values or test scores or any other nonnegative weights.
Again we assume here that the weights are convex coefficients and so act as proper probability-like mixture weights.
Ideally the combined FCM should also converge to or near the sampled underlying dynamical system with enough similar experts through some form of the law of large numbers \cite{taber2007quantization}.

FCM mixing contrasts with mixing finite \emph{augmented} Markov chains because then the mixed stochastic matrices of the Markov chains need not produce a stochastic matrix.
Markov chains can model causality with feedback but they weight their edges with transition probabilities. 
Their edge matrices need to be stochastic matrices whose rows consist of nonnegative numbers that sum to unity.
So Markov matrices cannot directly model negative causality or causal decrease as can the negative causal edge values in a FCM's causal edge matrix. 
The key point for knowledge representation is that FCM edge matrices are closed under general convex combination but the mixture of concept-similar stochastic matrices need not always produce a stochastic matrix. 

Markov-chain mixing is closed in the special case where all the mixed Markov chains use the exact same set of nodes.
The famous Birkhoff Theorem even shows that every doubly stochastic matrix is a convex mixture of permutation matrices. 
But mixing Markov chains with different sets of nodes does not give back a Markov chain in general.
Augmenting the chain matrices to include all $n$ nodes requires that for each stochastic matrix we add a zero-padded row and column for each missing node.
The resulting augmented $n$-by-$n$ matrices are no longer stochastic matrices.
And they do not mix to become a stochastic matrix or new Markov chain as we now illustrate.

Consider 3 Markov chains with 2 nodes each for a total of 3 distinct nodes $A$, $B$, and $C$.
Suppose the first Markov chains has nodes $A$ and $B$ and thus has a $2\times2$ stochastic matrix where $A$ and $B$ index both its 2 rows and 2 columns.
The second Markov chain has nodes $B$ and $C$.
The third Markov chain has nodes $A$ and $C$. 
Define their respective stochastic edge matrices $E_1$, $E_2$, and $E_3$ as
\begin{align}
    E_1 &= \begin{pmatrix}
        P_{AA} & P_{AB}\\
        P_{BA} & P_{BB}
    \end{pmatrix} = \begin{pmatrix}
        .2 & .8\\
        .7 & .3
    \end{pmatrix} \\E_2 &= \begin{pmatrix}
        P_{BB} & P_{BC}\\
        P_{CB} & P_{CC}
    \end{pmatrix} = \begin{pmatrix}
        .5 & .5\\
        .9 & .1
    \end{pmatrix} \\E_3 &= \begin{pmatrix}
        P_{AA} & P_{AC}\\
        P_{CA} & P_{CC}
    \end{pmatrix} = \begin{pmatrix}
        .4 & .6\\
        .2 & .8
    \end{pmatrix}
\end{align}
where $P_{ij}$ denotes the transition probability of going from state $i$ to state $j$. 

Now augment these 3 nonconforming $2\times2$ matrices into 3 respective conforming $3\times3$ edge matrices.
So add zero rows and zero columns for each missing node.
This gives the 3 $3\times3$ matrices $\Tilde{E}_1$, $\Tilde{E}_2$, and $\Tilde{E}_3$:
\begin{align}
    \Tilde{E}_1 = \begin{pmatrix}
.2 & .8 & 0\\
.7 & .3 & 0\\
0 & 0 & 0\end{pmatrix}, \:   \Tilde{E}_2 = \begin{pmatrix}
.5 & 0 & .5\\
0 & 0 & 0\\
.9 & 0 & .1\end{pmatrix}, \:  \Tilde{E}_3 = \begin{pmatrix}
0 & 0 & 0\\
0 & .4 & .6\\
0 & .2 & .8\end{pmatrix}.
\end{align}
Then mix these 3 augmented edge matrices with the respective convex mixing weights 0.3, 0.4, and 0.3  to give a non-stochastic matrix $E$:
\begin{align}
    E &= 0.3\begin{pmatrix}
.2 & .8 & 0\\
.7 & .3 & 0\\
0 & 0 & 0\end{pmatrix}+0.4\begin{pmatrix}
.5 & 0 & .5\\
0 & 0 & 0\\
.9 & 0 & .1\end{pmatrix}+0.3\begin{pmatrix}
0 & 0 & 0\\
0 & .4 & .6\\
0 & .2 & .8\end{pmatrix}\\
&=\begin{pmatrix}
.26 & .24 & .20\\
.21 & .21 & .18\\
.36 & .06 & .28\end{pmatrix}
\end{align}
No row in the combined matrix $E$ sums to unity.
So $E$ is not a stochastic matrix.
It is in this sense of combining augmented Markov chains that they are not closed under mixing.

Consider in the alternative mixing 3 nonconforming FCMs after augmenting their causal edge matrices.
The first FCM with nodes $C_1$, $C_2$, $C_3$, and $C_4$ has the edge matrix $E_1$. The second FCM with nodes $C_1$, $C_2$, $C_4$, and $C_5$ has the edge matrix $E_2$. The third FCM with nodes $C_2$, $C_3$, $C_4$, and $C_5$ has the edge matrix $E_3$. 
\begin{align}
    E_1 = \begin{pmatrix}
0 & 1 & 0 & 0\\
0 & 0 & -1 & 1\\
0 & 0 & 0 & 1\\
-1 & 0 & 0 & 0\end{pmatrix}, \:   E_2 = \begin{pmatrix}
0 & 1 & 0 & 1\\
0 & 0 & 1 & 0\\
-1 & 0 & 0 & 0\\
0 & 0 & 1 & 0\end{pmatrix}, \:   E_3 = \begin{pmatrix}
0 & -1 & 1 & 0\\
0 & 0 & 1 & 0\\
0 & 0 & 0 & 0\\
0 & 0 & 1 & 0\end{pmatrix}.
\end{align}

Augment these nonconforming $4\times4$ matrices with zero-rows and zero columns to give the respective $5\times5$ comfromable matrices $\Tilde{E}_1$, $\Tilde{E}_2$, and $\Tilde{E}_3$:
\begin{align}
    \Tilde{E}_1 = \begin{pmatrix}
0 & 1 & 0 & 0 & 0\\
0 & 0 & -1 & 1 & 0\\
0 & 0 & 0 & 1 & 0\\
-1 & 0 & 0 & 0 & 0\\
0 & 0 & 0 & 0 & 0\end{pmatrix}, \:   \Tilde{E}_2 = \begin{pmatrix}
0 & 1 & 0 & 0 & 1\\
0 & 0 & 0 & 1 & 0\\
0 & 0 & 0 & 0 & 0\\
-1 & 0 & 0 & 0 & 0\\
0 & 0 & 0 & 1 & 0\end{pmatrix}, \:   \Tilde{E}_3 = \begin{pmatrix}
0 & 0 & 0 & 0 & 0\\
0 & 0 & -1 & 1 & 0\\
0 & 0 & 0 & 1 & 0\\
0 & 0 & 0 & 0 & 0\\
0 & 0 & 0 & 1 & 0\end{pmatrix}.
\end{align}

These matrices then combine with respective convex coefficients 0.4, 0.3, and 0.3 to give edge matrix E of the FCM-mixture:
\begin{align}
    E &= 0.4\begin{pmatrix}
0 & 1 & 0 & 0 & 0\\
0 & 0 & -1 & 1 & 0\\
0 & 0 & 0 & 1 & 0\\
-1 & 0 & 0 & 0 & 0\\
0 & 0 & 0 & 0 & 0\end{pmatrix}+0.3\begin{pmatrix}
0 & 1 & 0 & 0 & 1\\
0 & 0 & 0 & 1 & 0\\
0 & 0 & 0 & 0 & 0\\
-1 & 0 & 0 & 0 & 0\\
0 & 0 & 0 & 1 & 0\end{pmatrix}+0.3\begin{pmatrix}
0 & 0 & 0 & 0 & 0\\
0 & 0 & -1 & 1 & 0\\
0 & 0 & 0 & 1 & 0\\
0 & 0 & 0 & 0 & 0\\
0 & 0 & 0 & 1 & 0\end{pmatrix}\\
&=\begin{pmatrix}
0 & .7 & 0 & 0 & .3\\
0 & 0 & - .7 & 1 & 0\\
0 & 0 & 0 & .7 & 0\\
- .7 & 0 & 0 & 0 & 0\\
0 & 0 & 0 & .6 & 0\end{pmatrix}
\end{align}
Every edge of the FCM-mixture is bipolar: $(E)_{ij} \in [-1,1]$. 
Mixing FCMs gives back FCMs.

FCM-mixing also contrasts Bayesian Belief Networks (BBNs). 
BBNs model causality with Directed Acyclic Graphs (DAGs) with conditional probability weighing the edges\cite{neuberg2003causality}. 
But conditional probability is not transitive and DAGs are not closed under mixing. 
Let $A$, $B$, and $C$ be three events. 
The conditional probability $P(C|A)$ is not always equal to the product $P(C|B)P(B|A)$. 
Intransitive conditional probability may not model transitive causality well. 
BBNs also assume joint probability distribution functions over their nodes. 
BBN inference needs NP-hard marginalization\cite{dagum1993approximating,russell2016artificial} through message-passing algorithms such as belief propagation\cite{yedidia2000generalized,murphy2002dynamic} or more general junction tree algorithm\cite{wainwright2008graphical}. 
This makes BBNs hard to scale. 
BBN learning from data is a NP-complete problem in general\cite{chickering1996learning,chickering2004large}.
Also: Convex combinations of DAGs are not always DAGs\cite{osoba2019beyond}. 
This prevents BBNs from mixing and does not allow knowledge combination.

\section{Phantom Nodes in FCMs} \label{Phantom Nodes}
\vspace{-6pt}
Augmented FCM matrices suggest a way to address the general problem of missing causal variables in a model.
An expert may not account for every causal variable in a dynamical system. 
Experts construct FCMs to model causal dynamical systems based on their understanding of the system. 
There is no way of knowing if there are causal variables in the system that the experts do not know about or forgot to account for. 
These causal variables do not appear as nodes on the FCMs but they are present in the system and causally affect the variables in the FCM. 

Phantom nodes are these hidden causal variables present in the system that the expert did not account for when constructing the FCM. 
The dynamical system feels their presence even though they do not appear on the FCM. 
This resembles how an amputee hallucinates phantom pain in their missing limb. 
Phantom nodes can augment the expert's FCM and approximate the limit cycles of the underlying dynamical system that the FCM models. 
The new edge $E^{PH}$ matrix for the augmented FCM pads the FCMs edge matrix $E$ with extra rows and columns corresponding to the phantom nodes. 
The estimates of edges between these phantom nodes and visible nodes may suggest to the expert what these hidden causal variables are. 

Figure~\ref{fig:phantom-dolphin-fcm} gives an example of a FCM with a phantom node. 
The expert did not account for the dolphins' ``{\footnotesize SURVIVAL~THREAT}" behavior. 
So node $C_4$ appears on the FCM as a phantom node. 
It is not visible but it still causally affects nodes $C_1$, $C_3$, and $C_5$. 

\begin{figure}[ht]
\centering
\includegraphics[width=0.55\textwidth]{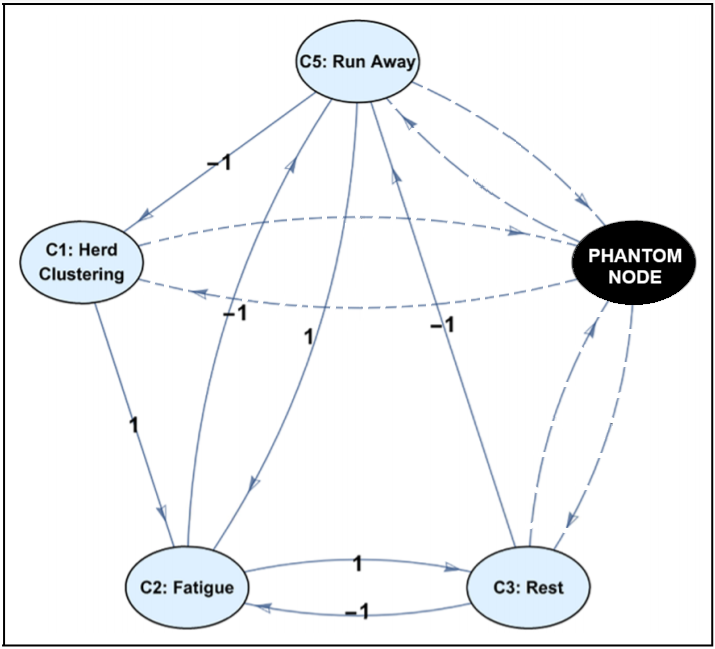}
\caption{The Dolphin FCM with a phantom node $C_4$ that corresponds to survival threats like sharks.
The domain expert or causal learning system does not observe $C_4$ or know the edges that correspond to $C_4$.}
\label{fig:phantom-dolphin-fcm}
\end{figure}

\vspace{-6pt}
\subsection{Effect of Phantom Nodes} \label{Phantom-effect}

Phantom nodes do not appear on the FCM but they do causally affect the nodes of the FCM. 
This may lead to limit cycles in the dynamical system that differ from those predicted by the FCM model. 
It may appear as if the FCM is hallucinating these limit cycles for some unknown reason. 
These differences in the limit cycles indicate the presence of phantom nodes. 

Figure~\ref{fig:phantom-effect} shows the effect of a phantom node. 
The actual dolphin behavior does not match the equilibrium of the expert's FCM. 
The expert's FCM is missing the node ``{\footnotesize SURVIVAL~THREAT}". 
It predicts that the dolphins will keep resting because it does not account for the presence of sharks. 
But sharks do causally affect the dolphin behavior as a phantom node. 
The dolphins go through a cycle of ``{\footnotesize HERD~CLUSTERING}", ``{\footnotesize FATIGUE}", \& ``{\footnotesize RUN~AWAY}"$\rightarrow$ ``{\footnotesize FATIGUE}" \& ``{\footnotesize REST}"$\rightarrow$ ``{\footnotesize REST}", \& ``{\footnotesize SURVIVAL~THREAT}"$\rightarrow$ ``{\footnotesize HERD~CLUSTERING}" \& ``{\footnotesize SURVIVAL~THREAT}"$\rightarrow$ ``{\footnotesize HERD~CLUSTERING}", ``{\footnotesize FATIGUE}", \& ``{\footnotesize RUN~AWAY}". 
The expert FCM cannot explain this behavior and might interpret it as a hallucination. 

\begin{figure}[ht]
\centering
\includegraphics[width=\textwidth]{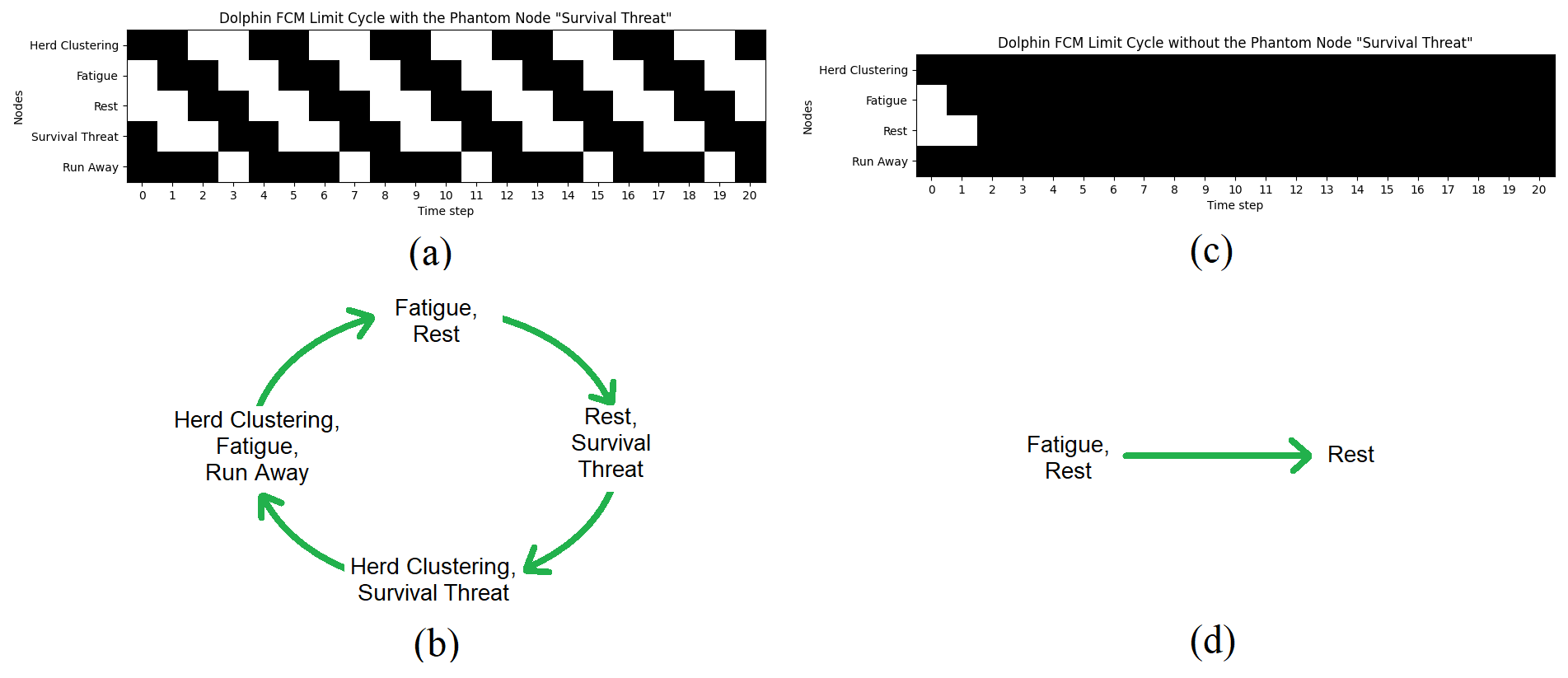}
\caption{Dolphin FCM limit cycles with and without phantom nodes. 
(a) Limit cycle of the Dolphin FCM with ``{\footnotesize SURVIVAL THREAT}" as a phantom node. 
The FCM starts at the initial state $\begin{pmatrix}0 & 1 & 1 & 0 & 0\end{pmatrix}$ and then goes through the cycle $\begin{pmatrix}0 & 1 & 1 & 0 & 0\end{pmatrix}\rightarrow\begin{pmatrix}0 & 0 & 1 & 1 & 0\end{pmatrix}\rightarrow\begin{pmatrix}1 & 0 & 0 & 1 & 0\end{pmatrix}\rightarrow\begin{pmatrix}1 & 1 & 0 & 1 & 0\end{pmatrix}\rightarrow\begin{pmatrix}0 & 1 & 1 & 0 & 0\end{pmatrix}$. 
(b) This limit cycle corresponds to the cycle of the dolphins as they rest from fatigue, face a survival threat while resting, cluster in a herd to evade the threat, and get tired again from running away in a herd.
(c) The limit cycle of the Dolphin FCM without the ``{\footnotesize SURVIVAL THREAT}" phantom node. 
The FCM starts at the same initial state $\begin{pmatrix}0 & 1 & 1 & 0\end{pmatrix}$. 
But it goes through the state $\begin{pmatrix}0 & 0 & 1 & 0\end{pmatrix}$ and then gets stuck at the fixed-point state $\begin{pmatrix}0 & 0 & 0 & 0\end{pmatrix}$. 
(d) This corresponds to the dolphins staying at rest because there is no threat. }
\label{fig:phantom-effect}
\end{figure}

\vspace{-6pt}
\subsection{Phantom Node Estimation} \label{Phantom-Estimation}

The equilibrium attractors characterize a dynamical system. 
FCMs that model the dynamical system should approximate its equilibrium behavior. 
The difference in the FCM's limit cycles and those of the dynamical system indicates the presence of phantom nodes. 
Phantom nodes can augment the FCM and approximate the dynamical system's limit cycles. 

The edge matrix $E^{PH}$ of this augmented FCM will have 4 blocks. The edge matrix $E$ denotes all the edges in the expert's FCM, the matrix $E^{OP}$ denotes all the edges going from observable nodes to the phantom nodes, the matrix $E^{PO}$ denotes all the edges going from the phantom nodes to the observable nodes, and the matrix $E^{PP}$ denotes all the edges between phantom nodes. 
Then the edge matrix $E^{PH}$ of the phantom-node-augmented FCM is
\begin{align}
    E^{PH} = \begin{pmatrix} E^{PP} & E^{PO} \\
                             E^{PO} & E      \end{pmatrix}
\end{align}

The phantom-node-augmented FCM trains by sampling from the limit cycles of the dynamical system. 
It compares its limit cycles to those of the dynamical system and finds the error between them. 
This error could simply be squared-error or it could be some kind of entropic error. 
A gradient-based learning algorithm can then estimate the edges between the FCM nodes and the phantom nodes by minimizing the error. 

Consider the $k$ steps $C(1)$, $C(2)$, $\dots$, $C(k)$ of the limit cycle of the causal dynamical system. 
Let $\Tilde{C(1)}$, $\Tilde{C(2)}$, $\dots$, $\Tilde{C(k)}$ denote the limit cycle of the FCM when augmented with the phantom nodes. 
The squared-error loss $L$ between the limit cycles is
\begin{align}
    L = \sum_{t=1}^k ||C(t)-\Tilde{C(t)}||^2. \label{eq:squared-error}
\end{align}
The edge matrix $E$ after $\tau$ epochs of training is 
\begin{align}
    E^{PH} (\tau+1) = E^{PH} (\tau) - \nabla_{E^{Ph}} L \label{eq:learning-phantom-nodes}
\end{align}
This may not learn the exact phantom node or the exact causal edge between the phantom node and the rest of the FCM. 
But it will learn phantom nodes and edges that produce approximately same limit cycles as the dynamical system. 

Dolphin FCM with 4 visible nodes $C_2$, $C_3$, $C_4$, and $C_5$ and 1 phantom node $C_1$ ran with 10,000 random initial conditions. 
The phantom-edge approximator trained by sampling the first 2 steps from the resulting limit cycles and estimated $E^{PH}$:
\begin{align}
E^{PH} = \begin{pmatrix}
0 & .6685 & .4392 & .0066 & .8296\\
 .6685 & 0 & 1 & 0 & -1\\
 .4392 & -1 & 0 & 1 & -1\\
 .0066 & 0 & -1 & 0 & 1\\
 .8296 & 1 & 0 & -1 & 0
\end{pmatrix}. \label{eq:learnt-phantom-node}
\end{align}
The edges do not match the edges on the original dolphin FCM but figure~\ref{fig:approx-limit-cycles} shows that the augmented FCM approximates the limit cycles of the target dynamical system. 

\begin{figure}[ht]
\centering
\includegraphics[width=0.8\textwidth]{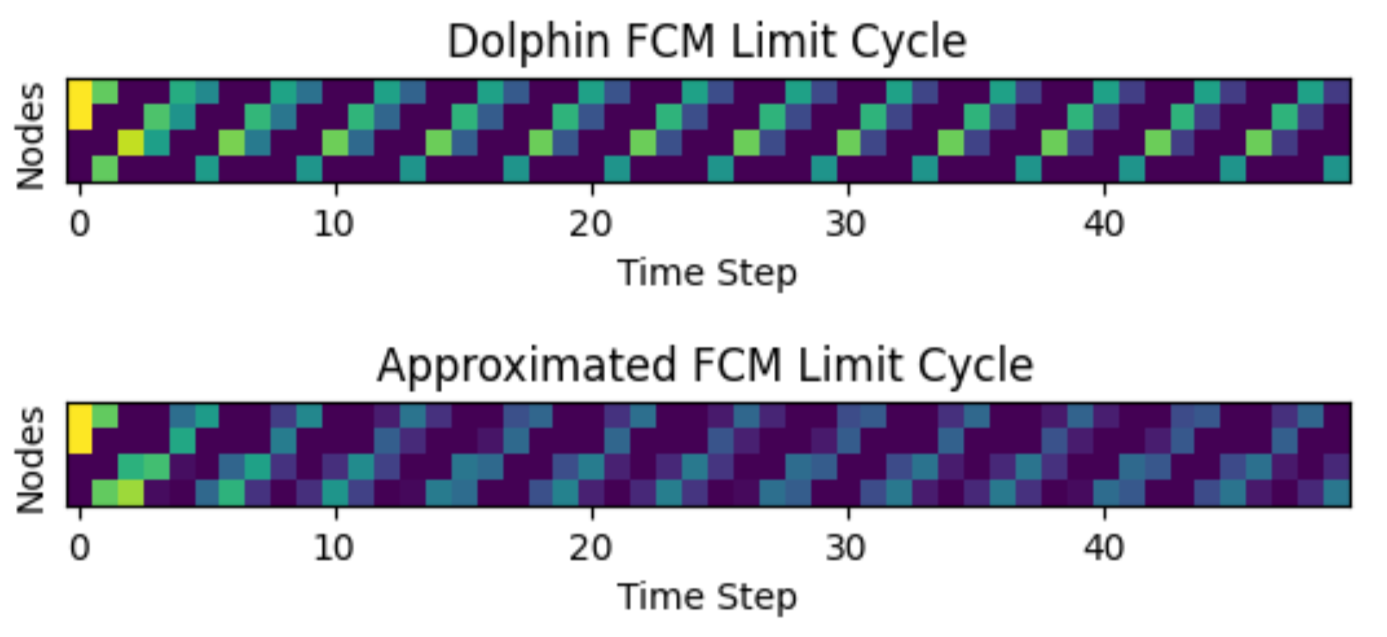}
\caption{The approximated limit cycles through phantom-node causal learning. 
The time step is along the $x$-axis. 
The 4 observable nodes lie along the $y$-axis. 
The color of the node represents its value. 
Bright color represents high value. 
Yellow nodes have value 1. 
Purple nodes have value 0. 
The initial state is $\begin{pmatrix}1 & 1 & 0 & 0\end{pmatrix}$ and converges to limit cycles in both FCMs. 
These approximated limit cycles are similar to the limit cycles of the Dolphin FCM. }
\label{fig:approx-limit-cycles}
\end{figure}

The learning algorithm may also use an entropic loss function
\begin{align}
    L = - \sum_{t = 1}^k \sum_{i = 1}^n \big[C_i(t)\ln C_i(t) + \big(1 - C_i(t)\big)\ln\big(1 - C_i(t)\big)\big]
\end{align}

\section{Mixture Combination of Phantom-augmented FCMs} \label{phantom-combination}
\vspace{-6pt}
Multiple experts may model the same causal dynamical system differently. 
Some experts may account for causal variables that the other experts did not account for.
There will in general also be causal variables that none of the experts accounted for. 
Then the mixed or combined FCM from all the experts will likely still not itself account for all the relevant causal variables in the system. 
But accounting for phantom nodes in the individual FCMs before combining them may better approximate the dynamical system. 
We now explore this divide-and-conquer-then-mix strategy for estimating phantom nodes.

Consider $m$ expert FCMs that model the same causal dynamical system. 
Let $E_i$ denote the edge matrix of the $i$th FCM. 
The phantom nodes of the $i$th FCM augment it with their associated matrices $E^{PP}_i$, $E^{PO}_i$, and $E^{OP}_i$ to give
\begin{align}
    E^{PH}_i = \begin{pmatrix} E^{PP}_i & E^{PO}_i \\
                             E^{PO}_i & E_i      \end{pmatrix}.
\end{align}
Equations~\ref{eq:squared-error} and \ref{eq:learning-phantom-nodes} can learn this phantom-node-augmented edge matrix $E^{PH}_i$. 
We then pad $E^{PH}_i$ with zero-rows and zero-columns corresponding to the nodes present in the other $E^{PH}_j$ matrices but absent in $E^{PH}_i$ to give the new edge matrix $\Tilde{E}^{PH}_i$. 
This makes all $\Tilde{E}^{PH}_i$s conformable for matrix addition. 
Then equation~\ref{eq:fcm-combination} combines these augmented-FCMs to give the edge matrix $E$ of the combined FCM:
\begin{align}
    E = \sum_{i=1}^m w_i\Tilde{E}^{PH}_i
\end{align}

This mixture gives a way to represent and combine multiple hidden causal factors in parallel.

\section{Experimental Results} \label{experiments}
\vspace{-6pt}
Consider the 5-node dolphin FCM as the causal dynamical system that the experts want to model. 
Three experts each come up with a 4-node FCM to explain the dolphin behavior. The first expert does not account for node $C_1$ corresponding to ``{\footnotesize HERD~CLUSTERING}", the second expert forgets to account for $C_2$ corresponding to ``{\footnotesize FATIGUE}", and the third experts does not account for the node $C_4$ corresponding to ``{\footnotesize SURVIVAL~THREAT}". All three FCMs run from 10,000 random initial conditions. The corresponding edge matrices are
\begin{align}
    E_1 = \begin{pmatrix}
0 & 1 & 0 & -1\\
-1 & 0 & 1 & -1\\
0 & -1 & 0 & 1\\
1 & 0 & -1 & 0\end{pmatrix}, \:   E_2 = \begin{pmatrix}
0 & 0 & -1 & 0\\
0 & 0 & 1 & -1\\
1 & -1 & 0 & 1\\
-1 & 0 & -1 & 0\end{pmatrix}, \:  E_3 = \begin{pmatrix}
0 & 1 & 0 & 0\\
0 & 0 & 1 & -1\\
0 & -1 & 0 & -1\\
-1 & 1 & 0 & 0\end{pmatrix}.
\end{align}

We then augment each 4-node FCM with one phantom node each. 
The augmented FCMs then train by sampling the first two steps of the limit cycles from the 10,000 runs. 
This gives 3 5-node FCMs with edge matrices $E^{PH}_1$, $E^{PH}_2$, and $E^{PH}_3$:
\begin{align}
E^{PH}_1 = \begin{pmatrix}
0 & .6685 & .4392 & .0066 & .8296\\
 .6685 & 0 & 1 & 0 & -1\\
 .4392 & -1 & 0 & 1 & -1\\
 .0066 & 0 & -1 & 0 & 1\\
 .8296 & 1 & 0 & -1 & 0
\end{pmatrix},
\end{align}
\begin{align}
E^{PH}_2 = \begin{pmatrix}
0 & .8601 & .6264 & .9446 & .4425\\
 .8601 & 0 & 0 & -1 & 0\\
 .6264 & 0 & 0 & 1 & -1\\
 .9446 & 1 & -1 & 0 & 1\\
 .4425 & -1 & 0 & -1 & 0
\end{pmatrix},
\end{align}
\begin{align}
E^{PH}_3 = \begin{pmatrix}
0 & .7535 & .4866 & .8142 & .0701\\
 .7535 & 0 & 1 & 0 & 0\\
 .4866 & 0 & 0 & 1 & -1\\
 .8142 & 0 & -1 & 0 & -1\\
 .0701 & -1 & 1 & 0 & 0
\end{pmatrix}.
\end{align}
The combined FCM will have 5 observable nodes and 3 phantom nodes. 
So we pad the $5\times5$ $E^{PH}_i$ matrices with zero-rows and zero-columns to get three $8\times8$ matrices $\Tilde{E}^{PH}_i$:
\begin{align}
\Tilde{E}^{PH}_1 = \text{\footnotesize $\begin{pmatrix}
0 & 0 & 0 & 0 & .6685 & .4392 & .0066 & .8296\\
0 & 0 & 0 & 0 & 0 & 0 & 0 & 0 \\
0 & 0 & 0 & 0 & 0 & 0 & 0 & 0 \\
0 & 0 & 0 & 0 & 0 & 0 & 0 & 0 \\
 .6685 & 0 & 0 & 0 & 0 & 1 & 0 & -1\\
 .4392 & 0 & 0 & 0 & -1 & 0 & 1 & -1\\
 .0066 & 0 & 0 & 0 & 0 & -1 & 0 & 1\\
 .8296 & 0 & 0 & 0 & 1 & 0 & -1 & 0
\end{pmatrix}$},
\end{align}
\begin{align}
\Tilde{E}^{PH}_2 = \text{\footnotesize $\begin{pmatrix}
0 & 0 & 0 & 0 & 0 & 0 & 0 & 0 \\
0 & 0 & 0 & .8601 & 0 & .6264 & .9446 & .4425\\
0 & 0 & 0 & 0 & 0 & 0 & 0 & 0 \\
0 & .8601 & 0 & 0 & 0 & 0 & -1 & 0\\
0 & 0 & 0 & 0 & 0 & 0 & 0 & 0 \\
0 & .6264 & 0 & 0 & 0 & 0 & 1 & -1\\
0 & .9446 & 0 & 1 & 0 & -1 & 0 & 1\\
0 & .4425 & 0 & -1 & 0 & 0 & -1 & 0
\end{pmatrix}$},
\end{align}
\begin{align}
\Tilde{E}^{PH}_3 = \text{\footnotesize $\begin{pmatrix}
0 & 0 & 0 & 0 & 0 & 0 & 0 & 0 \\
0 & 0 & 0 & 0 & 0 & 0 & 0 & 0 \\
0 & 0 & 0 & .7535 & .4866 & .8142 & 0 & .0701\\
0 & 0 & .7535 & 0 & 1 & 0  & 0 & 0\\
0 & 0 & .4866 & 0 & 0 & 1  & 0 & -1\\
0 & 0 & .8142 & 0 & -1 & 0  & 0 & -1\\
0 & 0 & 0 & 0 & 0 & 0 & 0 & 0 \\
0 & 0 & .0701 & -1 & 1 & 0  & 0 & 0
\end{pmatrix}$}.
\end{align}

These 8-node FCMs then mix through convex combination to give the edge matrix $E$ of the combined FCM. 
We weight the experts equally so $w_1 = w_2 = w_3 = 1/3$. 
So 
\begin{align}
E =\text{\footnotesize $\begin{pmatrix}
0 & 0 & 0 & 0 & .22 & .01 & .00 & .28 \\
0 & 0 & 0 & .29 & 0 & .21 & .31 & .15 \\
0 & 0 & 0 & .25 & .16 & .27 & 0 & .02 \\
0 & .29 & .25 & 0 & .67 & 0 & - .67 & 0\\
 .22 & 0 & .16 & 0 & 0 & .67 & 0 & - .67\\
 .01 & .21 & .27 & 0 & - .67 & 0 & .67 & -1\\
 .00 & .31 & 0 & .67 & 0 & - .67 & 0 & .67\\
 .28 & .15 & .02 & - .67 & .67 & 0 & - .67 & 0
\end{pmatrix}$}
\end{align}

This FCM mixture predicts limit cycles that are close to the Limit cycles of the dolphin FCM. 
Figure~\ref{fig:Augmented-FCM-Mixture-limit-cycles} shows an instance of limit cycle approximation. 

\begin{figure}[ht]
\centering
\includegraphics[width= .8\textwidth]{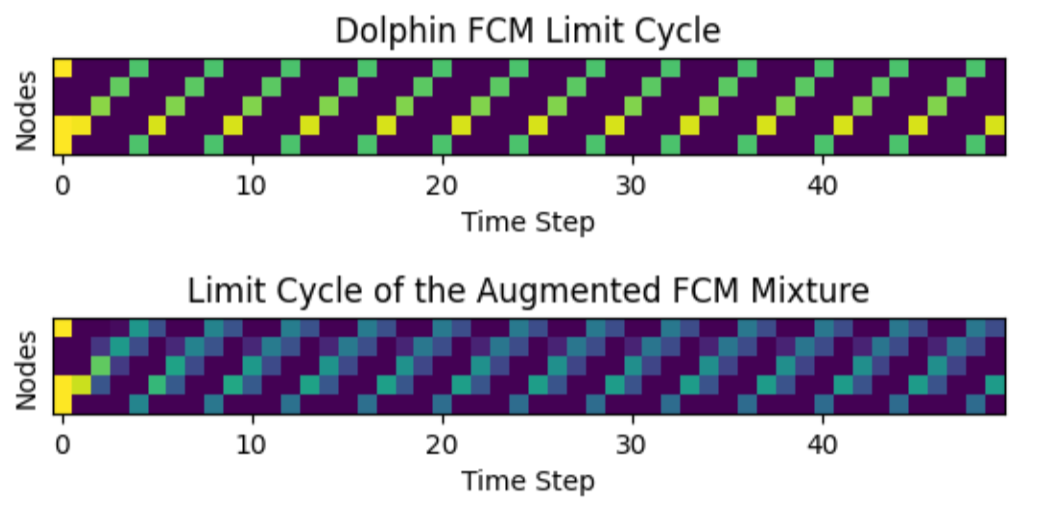}
\caption{The approximated limit cycles through phantom-node-augmented FCM mixture. 
The time step is along the $x$-axis. 
The 5 observable nodes lie along the $y$-axis. 
The color of the node represents its value. 
Bright color represents high value. 
Yellow nodes have value 1. 
Purple nodes have value 0. 
The FCM mixture combines 3 FCMs with 4 observable nodes and 1 phantom node each. 
The initial state is $\begin{pmatrix}1 & 0 & 0 & 1 & 1\end{pmatrix}$ and converges to limit cycles in both FCMs. 
These approximated limit cycles are similar to the limit cycles of the Dolphin FCM. }
\label{fig:Augmented-FCM-Mixture-limit-cycles}
\end{figure}

The augmented-FCM mixture approximates the dolphin limit cycles accurately even when a component FCM does not. 
The third FCM $E^{PH}_3$ does not learn its phantom node $C_4$ perfectly. 
Figure~\ref{fig:Augmented-FCM-wrong-limit-cycles} shows that it converges to a fixed point when it should be converging to a limit cycle. 
But the other phantom nodes in the FCM-mixture counter its causal effect such that the mixture correctly predicts the dolphin limit cycles. 
The performance improves with the number of components in this big-knowledge mixture. 

\begin{figure}[ht]
\centering
\includegraphics[width=0.8\textwidth]{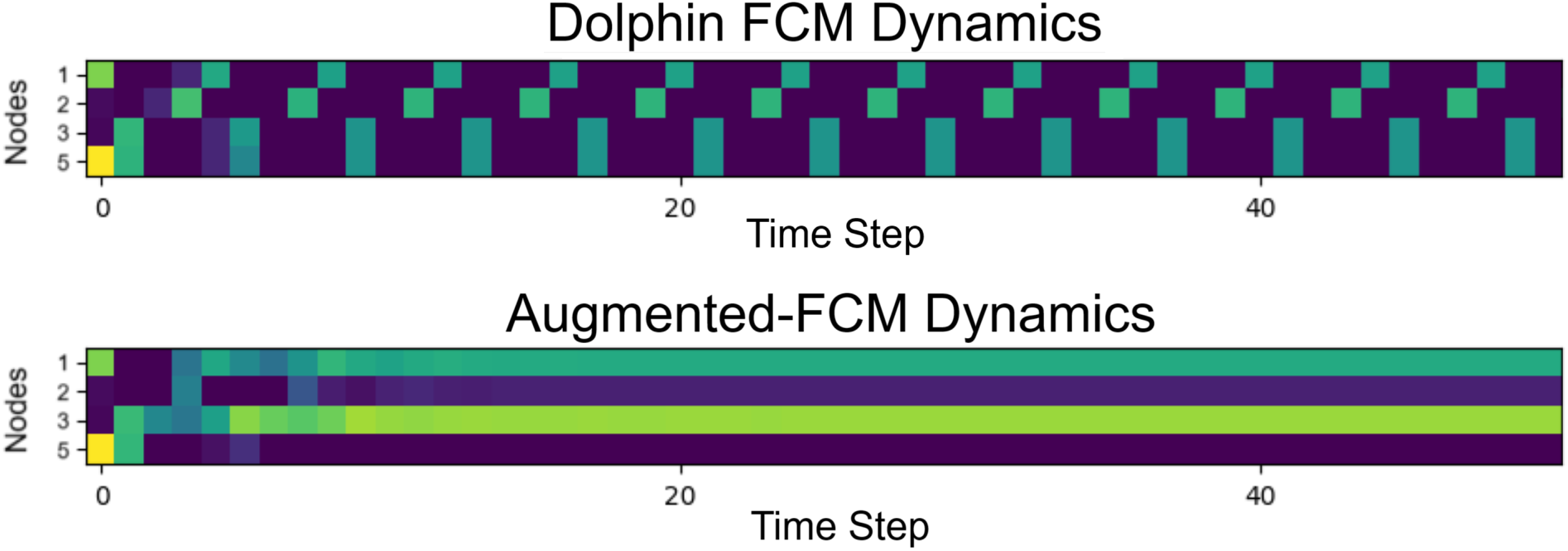}
\caption{The limit cycles with phantom node $C_4$. 
The time step is along the $x$-axis. 
The 4 observable nodes lie along the $y$-axis. 
The color of the node represents its value. 
Bright color represents high value. 
Yellow nodes have value 1. 
Purple nodes have value 0. 
The FCM $E^{PH}_3$ learns the wrong phantom node $C_4$ and fails to approximate the dolphin limit cycles. 
It converges to a fixed point instead of the dolphin FCM's limit cycle with the same initial state $\begin{pmatrix}1 & 0 & 0 & 1\end{pmatrix}$.}
\label{fig:Augmented-FCM-wrong-limit-cycles}
\end{figure}

\section{Conclusion}
\vspace{-6pt}
Estimating missing or phantom nodes in a causal model remains a hard problem because at root the task is to estimate the unknown unknowns that affect dynamical equilibria.
The direct estimate of a single phantom node in a feedback fuzzy cognitive map can be computationally heavy.  
The computational burden greatly increases with the number of phantom nodes.
Adapting mixed individual FCMs where each has a single estimated phantom node is comparatively tractable when the user knows at least the rough form of some of the equilibrium limit cycles or fixed-point attractors. 
Future algorithms can extend these techniques to multiple phantom nodes per expert and to multivariable feedback systems with richer equilibria and perhaps even chaotic equilibria.

\bibliographystyle{splncs04}
\bibliography{DMBD2024-Phantom-Node-Mixture-31Dec2024}

\end{document}